\documentclass[journal]{IEEEtran}
\usepackage{amsmath,amsfonts}
\usepackage{algorithmic}
\usepackage{algorithm}
\usepackage{array}
\usepackage[caption=false,font=normalsize,labelfont=sf,textfont=sf]{subfig}
\usepackage{textcomp}
\usepackage{stfloats}
\usepackage{url}
\usepackage{verbatim}
\usepackage{graphicx}
\usepackage{cite}
\usepackage{bibentry}
\usepackage{multirow}
\usepackage{color}
\usepackage{newfloat}
\usepackage{listings}
\usepackage{amsmath}
\usepackage{caption}
\usepackage{times} 
\usepackage{helvet}  
\usepackage{courier}  

\begin{document}

\title{SGCE-Font: Skeleton Guided Channel Expansion for Chinese Font Generation}

\author{Jie Zhou, Yefei Wang, Yiyang Yuan, Qing Huang, Jinshan Zeng$^\dag$
\thanks{Jie Zhou, Yefei Wang, Yiyang Yuan, Qing Huang and Jinshan Zeng are with the School of Computer and Information Engineering, Nanchang, 330022, China.
Email: 202041600009@jxnu.edu.cn (J. Zhou), fei@jxnu.edu.cn (Y. Wang), yiyangyuan218@163.com (Y.Yang), qh@jxnu.edu.cn (Q. Huang), jinshanzeng@jxnu.edu.cn (J. Zeng). 
$\dag$ Corresponding author: \textit{Jinshan Zeng}.}
}

\maketitle

\begin{abstract}
The automatic generation of Chinese fonts is an important problem involved in many applications. The predominated methods for the Chinese font generation are based on the deep generative models, especially the generative adversarial networks (GANs). However, existing GAN-based methods (say, CycleGAN) for the Chinese font generation usually suffer from the mode collapse issue, mainly due to the lack of effective guidance information. This paper proposes a novel information guidance module called the \textit{skeleton guided channel expansion (SGCE)} module for the Chinese font generation through integrating the skeleton information into the generator with the channel expansion way, motivated by the observation that the skeleton embodies both local and global structure information of Chinese characters. We conduct extensive experiments to show the effectiveness of the proposed module. Numerical results show that the mode collapse issue suffered by the known CycleGAN can be effectively alleviated by equipping with the proposed SGCE module, and the CycleGAN equipped with SGCE outperforms the state-of-the-art models in terms of four important evaluation metrics and visualization quality. Besides CycleGAN, we also show that the suggested SGCE module can be adapted to other models for Chinese font generation as a plug-and-play module to further improve their performance.
\end{abstract}


\section{Introduction}
\IEEEPARstart{T}{he} generation of Chinese fonts has attracted amounts of attention in recent years due to its wide range of applications \cite{zhang2019calligraphy,wang2022aesthetic,luo2022slogan}. With the development of deep learning, in particular the deep generative models such as the generative adversarial networks (GANs) \cite{goodfellow2014generative} and variational auto-encoder (VAE) \cite{Kingma2014-VAE}, the kind of deep generative model based methods has become the mainstream due to their impressive performance \cite{xu2005automatic,tian2017zi2zi,Yamada2017The,Goto2021Learning,park2021few,qin2022disentangled}. Existing deep generative models for Chinese font generation can be generally divided into two categories, that is, supervised models \cite{lyu2017auto,jiang2017dcfont,tian2017zi2zi,lei2018learning,chang2018chinese,https://doi.org/10.48550/arxiv.2005.12500,https://doi.org/10.48550/arxiv.2204.10484} and unsupervised models \cite{zhang2019calligraphy,qin2022disentangled,park2021few,park2021multiple, hassan2021unpaired}, where the kind of supervised models is mainly based on the paired data (i.e., there is a one-to-one correspondence between characters in the source and target font domains), and the kind of unsupervised models is mainly based on the unpaired data (without requiring a one-to-one correspondence between the source and target font domains).

In the early work \cite{lyu2017auto} of supervised models, the authors suggested an auto-encoder guided GAN model for the Chinese calligraphy synthesis based on the paired data, regarding the Chinese font generation problem as an image-to-image translation problem. Motivated by this perspective, many models have been developed for the Chinese font generation in recent years \cite{tian2017zi2zi,https://doi.org/10.48550/arxiv.2005.12500,https://doi.org/10.48550/arxiv.2204.10484}. In \cite{tian2017zi2zi}, the authors adopted the Pix2Pix model \cite{isola2017image} to the generation of Chinese fonts. In \cite{https://doi.org/10.48550/arxiv.2005.12500}, an effective GAN model dubbed \textit{CalliGAN} was proposed for the Chinese font generation by exploiting certain component information of Chinese characters. The recent paper \cite{https://doi.org/10.48550/arxiv.2204.10484} proposed a GAN-based image translation model for the synthesis of Chinese brush handwriting font by integrating the skeleton information in the expense of increasing model complexity of the deep neural networks. Despite the impressive performance of this kind of supervised models \cite{lyu2017auto,tian2017zi2zi,https://doi.org/10.48550/arxiv.2005.12500,https://doi.org/10.48550/arxiv.2204.10484}, the collection of a large amount of paired samples is generally overhead, and even unrealistic for some font generation problems like the generation of ancient calligraphy fonts \cite{qian2007towards,chen2011chinese}.

\begin{figure*}[t]
\begin{minipage}[b]{0.99\linewidth} 
\centering
\includegraphics*[scale=0.3]{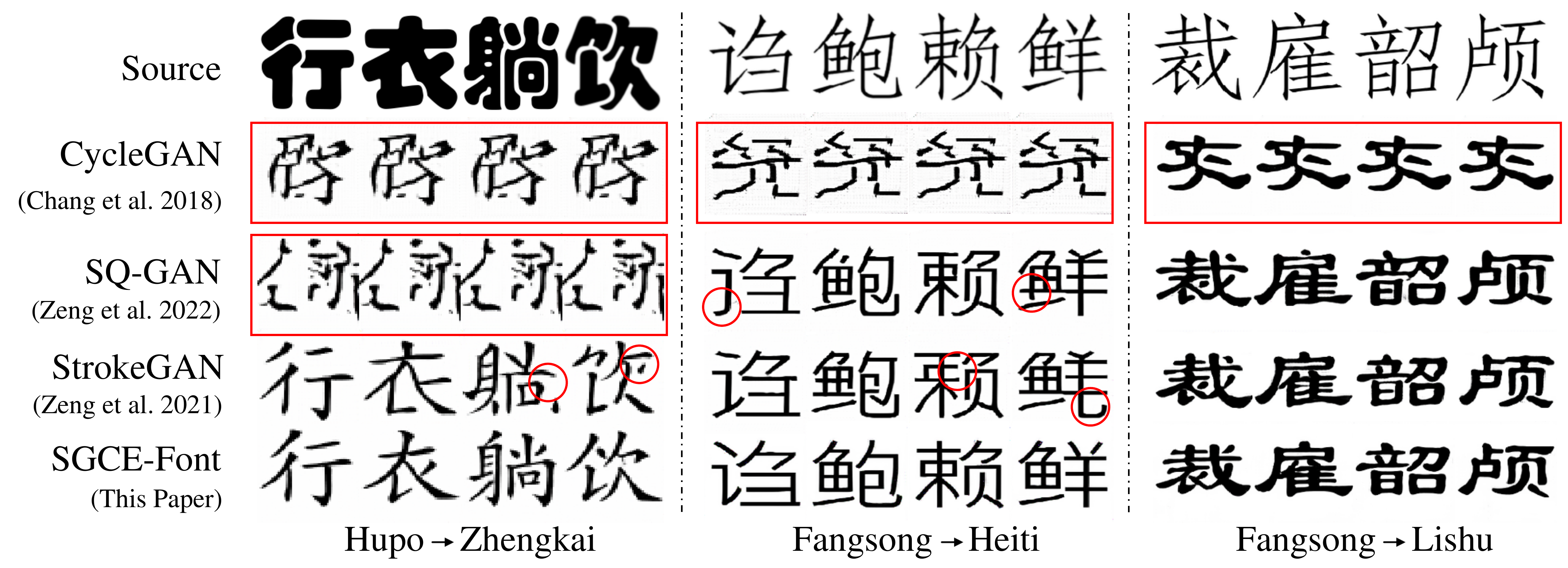}
\end{minipage}
\hfill
\flushleft
\caption{Alleviating mode collapse by the proposed SGCE module over three generation tasks. 
From left to right of the tasks: Hupo font to Zhengkai font, Fangsong to Heiti, Fansgong to Lishu. 
The second to the last rows indicate different methods:
CycleGAN \cite{chang2018generating} encounters mode collapse over these three generation tasks. CycleGAN with square-block transformations (SQ-GAN) \cite{zeng2022SQ-GAN} also encounters mode collapse in the task \{Hupo$\rightarrow$Zhengkai\}. CycleGAN with stroke encodings (StrokeGAN) \cite{zeng2021strokegan} 
generates characters with some flaws.  The CycleGAN equipped with the proposed SGCE module (SGCE-Font) can effectively address the mode collapse issue and generate very realistic Chinese characters.}
\label{fig:modecollapse}
\end{figure*}

To tackle this issue, many unsupervised models based on unpaired data have been suggested in the literature \cite{chang2018generating,jiang2019scfont,lin2020chinese,xie2021dg}. In the seminal work \cite{chang2018generating} of unsupervised models, the authors adopted the well-known CycleGAN model \cite{zhu2017unpaired} to the generation of Chinese fonts based on the unpaired data. In \cite{li2019improving}, the authors proposed an effective model using graph matching for the calligraphy character generation. In \cite{jiang2019scfont}, a structure-guided deep generative model dubbed \textit{SCFont} was suggested for the generation of Chinese fonts, 
by integrating the domain knowledge of Chinese characters such as writing trajectory and skeleton into the generation. \cite{xie2021dg} proposed a novel deformable generative model called \textit{DG-Font} for the generation of Chinese font based on unpaired data. Although the effectiveness of these unsupervised models has been demonstrated in the literature, the kind of unsupervised models (say, CycleGAN \cite{chang2018generating}) usually suffer from the well-known mode collapse issue \cite{goodfellow2014generative}, which results in poor performance, mainly due to the lack of effective guidance information. As shown in the second row of Figure \ref{fig:modecollapse}, mode collapse happens for CycleGAN in three concerned generation tasks, and in this case, CycleGAN fails to yield the correct Chinese characters.

\begin{figure}[!h]
\begin{minipage}[b]{0.99\linewidth}
\centering
\includegraphics[scale=0.32]{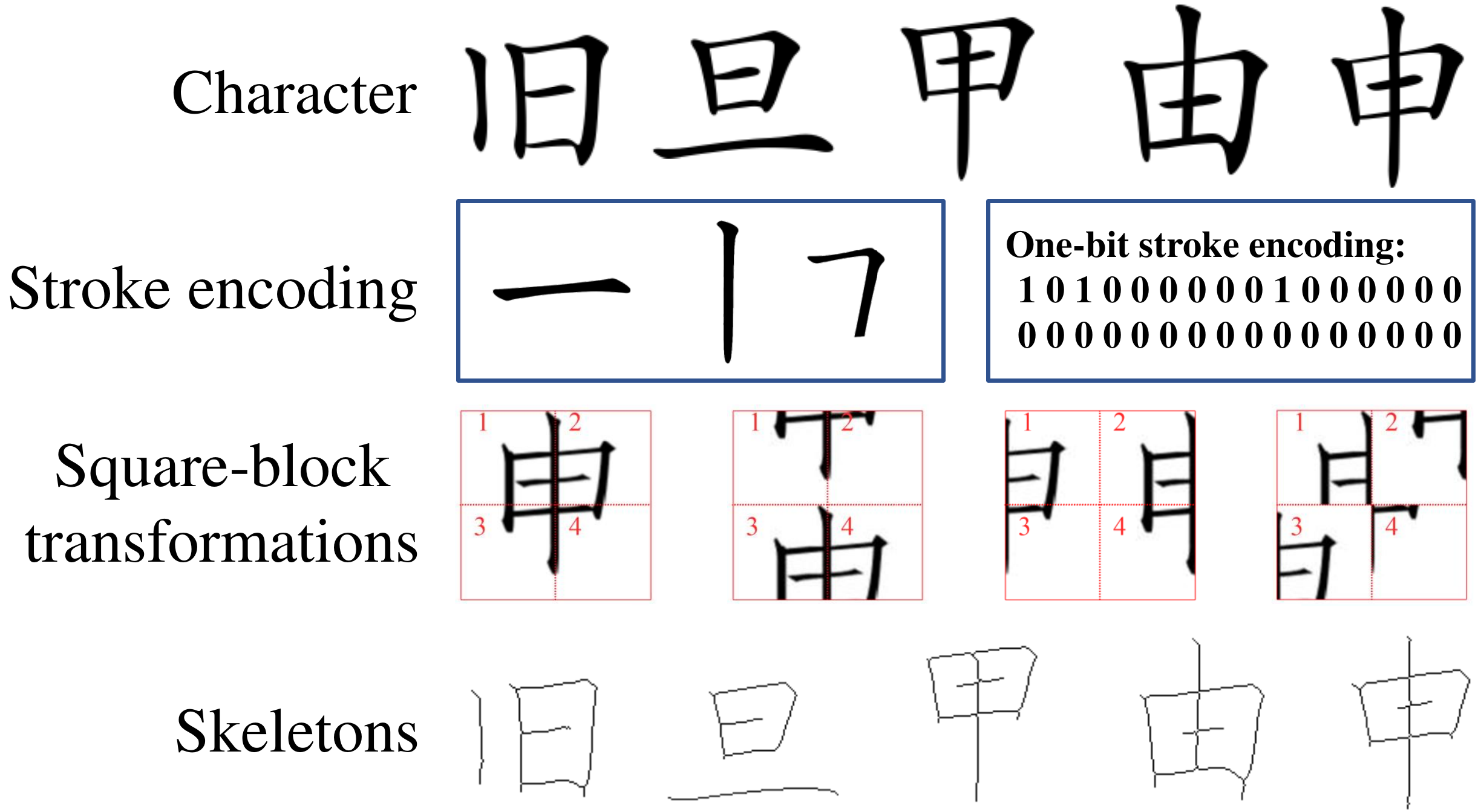}
\end{minipage}
\hfill
\caption{Examples of Chinese characters and their stroke encodings, square-block transformations and skeletons.
 }
\label{fig:skeleton_stroke}
\end{figure}

\begin{figure}[!t]
\begin{minipage}[b]{0.99\linewidth}
\centering
\includegraphics*[scale=0.22]{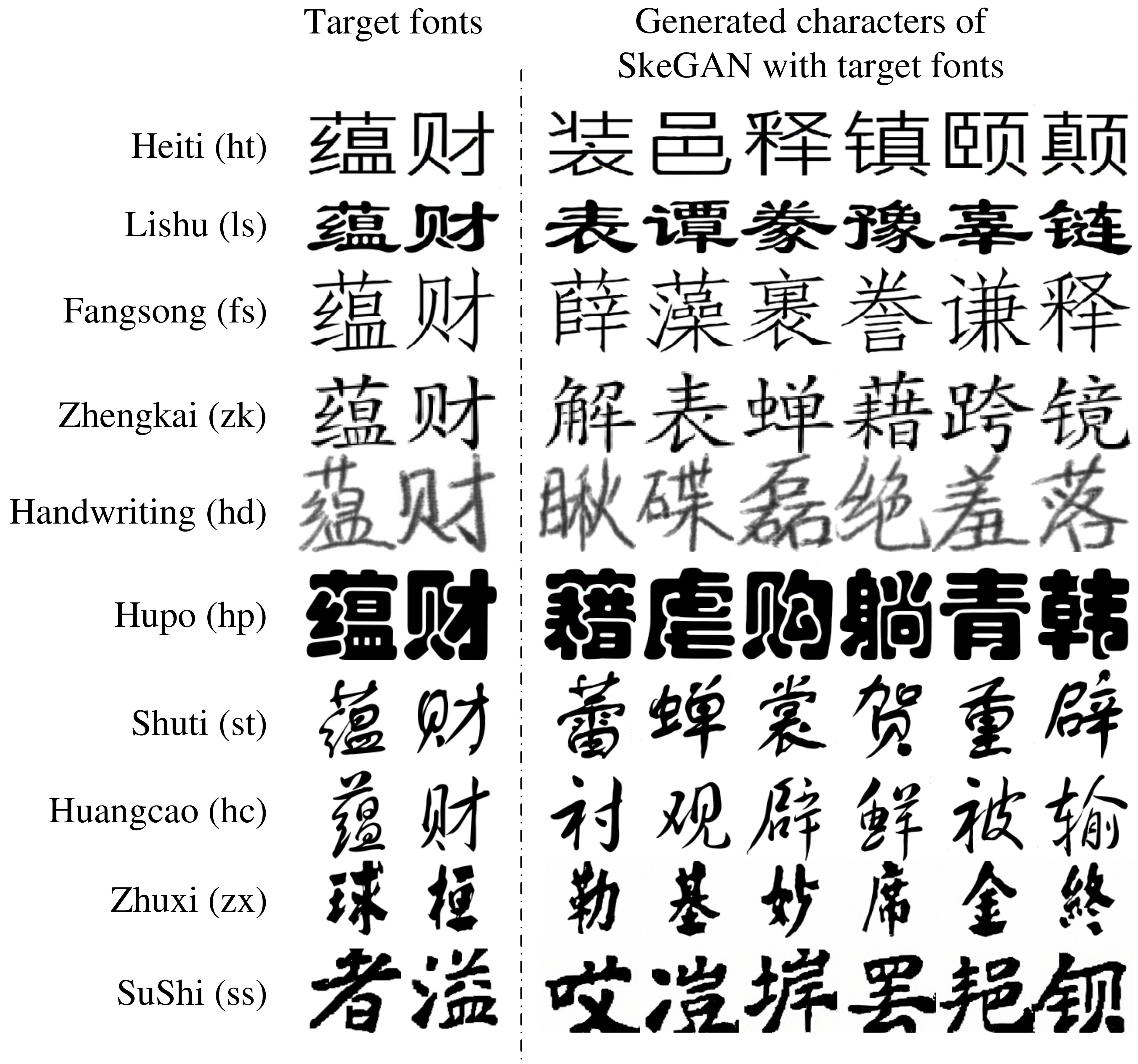}
\end{minipage}
\hfill
\caption{Generated characters of the proposed SGCE-Font model with the known CycleGAN as the base model for the generation of concerned ten Chinese fonts.
}
\label{fig:generated-characters}
\end{figure}

To reduce the mode collapse, several effective Chinese font generation models have been recently suggested in the literature \cite{zeng2021strokegan,zeng2022SQ-GAN} through integrating certain kinds of component information of Chinese characters.
In \cite{zeng2021strokegan}, the authors incorporated the stroke information into CycleGAN to reduce the mode collapse (there dubbed \textit{StrokeGAN}), where certain one-bit stroke encoding was introduced to represent the stroke information as shown in the second row of Figure \ref{fig:skeleton_stroke}. In \cite{zeng2022SQ-GAN}, the authors proposed a square-block geometric transformation based self-supervised scheme for CycleGAN to capture certain spatial structure information of Chinese characters, inspired by the observation that character structures are closely related to these square-block transformations. As shown in the third row of Figure \ref{fig:skeleton_stroke}, these four square-block transformations were used in \cite{zeng2022SQ-GAN} to represent the single-component, left-right, up-bottom, and semi-encircling structures, respectively. It can be observed from the third and fourth rows of Figure \ref{fig:modecollapse} that both the one-bit stroke encoding and square-block transformations are generally insufficient to thoroughly address the mode collapse issue, since there are some Chinese characters with both the same one-bit stroke encodings and character structures as shown in Figure \ref{fig:skeleton_stroke}. Moreover, the guidance information used in both StrokeGAN \cite{zeng2021strokegan} and SQ-GAN \cite{zeng2022SQ-GAN} was imposed on the discriminator, leading to certain implicit guidance during the generation.

In this paper, we focus on developing the effective information guidance scheme for the Chinese font generation. In particular, we aim to answer the following two questions:
\begin{enumerate}
    \item [(Q1)] What kind of guidance information is more effective for the Chinese font generation? 
    \item[(Q2)] How to use the guidance information more effectively for the Chinese font generation?
\end{enumerate}

For (Q1), we suggest using the skeletons of Chinese characters as the guidance information for the Chinese font generation, motivated by the observation that the skeleton can embody both local and global structure information of Chinese characters as well as their semantic information, as shown in Figure \ref{fig:skeleton_stroke}. From Figure \ref{fig:skeleton_stroke}, the stroke encodings used in StrokeGAN \cite{zeng2021strokegan} and square-block transformations used in SQ-GAN \cite{zeng2022SQ-GAN} can only reflect either local or global structure information of Chinese characters and in particular cannot provide the sole representation of a Chinese character, while the skeleton can provide both local and global structure information as well as the sole representation of a Chinese character. Thus, the skeleton should be more effective than both the stroke encoding and square-block transformation for the Chinese font generation. For (Q2), we directly use the skeleton information as a part of input of the generator through a channel expansion way, instead of imposing it on the discriminator. Thus, the skeleton should provide more information and stronger guidance for the generator than existing implicit ways of imposing guidance information on the discriminator.

\begin{figure}[!t]
\begin{minipage}[b]{0.99\linewidth}
\centering
\includegraphics[scale=0.5]{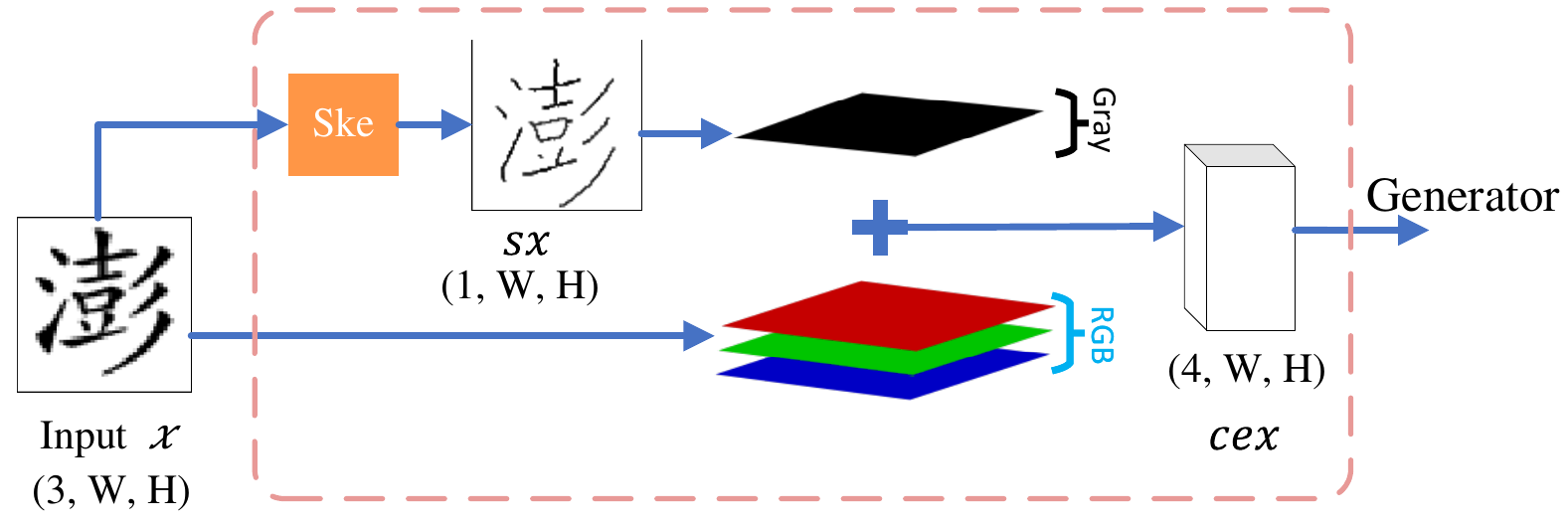}
\end{minipage}
\hfill
\caption{Skeleton guided channel expansion (SGCE) module for Chinese font generation.
}
\label{fig:SGCE}
\end{figure}

\begin{figure*}[!t]
\begin{minipage}[b]{0.99\linewidth}
\centering
\includegraphics[scale=0.70]{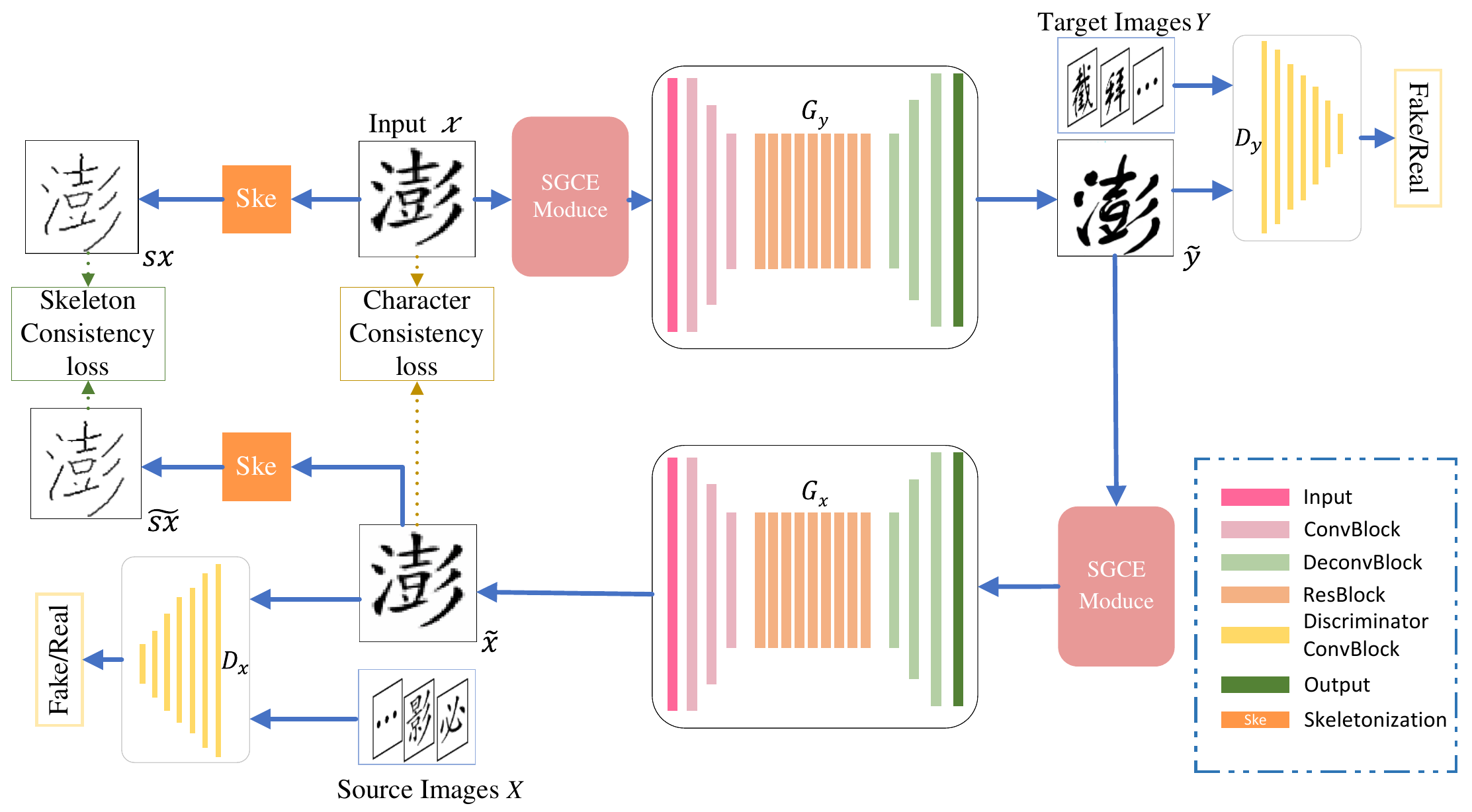}
\end{minipage}
\hfill
\caption{The pipeline of the proposed SGCE-Font model for Chinese font generation, where we take CycleGAN \cite{chang2018generating} as the base model as an example.
}
\label{fig:SGCE-Font}
\end{figure*}
The major contributions of this paper can be summarized as follows:
\begin{enumerate}
    \item[(1)] This paper proposes a novel information guidance module called the \textit{skeleton guided channel expansion} (SGCE) for the Chinese font generation through integrating the skeleton information into the generator by a channel expansion way. The proposed module can provide both local and global structure information as the guidance information for the Chinese font generation while does not increase any additional model complexity. Thus, the proposed module can effectively alleviate the mode collapse and improve the quality of the generated Chinese characters, and can be used as a plug-and-play module for Chinese font generation models without increasing model complexity.
    \item[(2)] Extensive experiments are conducted to demonstrate the effectiveness of the proposed module, in particular its superiority on reducing mode collapse and enhancing the generation performance. The outperformance of the proposed SGCE-Font model (i.e., equipping the proposed SGCE module into the known CycleGAN model) is also shown by numerous experiments under quantitative comparisons with the state-of-the-art models. Some generated characters of the proposed model are presented in Figure \ref{fig:generated-characters}. It can be observed that the proposed model can generate very realistic Chinese characters over these ten concerned Chinese fonts.
    \item[(3)] Besides CycleGAN, we also show that the proposed module can be easily adapted to other existing Chinese font generation models and further enhance their performance.
\end{enumerate}

The rest of this paper is organized as follows. In Section \ref{sc:releated work}, we present some related work. In Section \ref{sc:proposed model}, we describe the proposed module in detail. In Section \ref{sc:experiments}, we provide amounts of experiments to demonstrate the effectiveness of the proposed module. We conclude this paper in the final section.

\section{Related Work}
\label{sc:releated work}

As shown in Figure \ref{fig:skeleton_stroke}, the components of Chinese characters such as strokes and skeletons directly reflect certain structure information of Chinese characters. Motivated by this, many kinds of component information of Chinese characters have been widely used in the literature \cite{lian2012automatic,liu2012automatic,lin2015complete,lin2019font,wen2021handwritten,https://doi.org/10.48550/arxiv.2204.10484,jiang2019scfont,gao2020gan,qin2022disentangled,Y2021Calligraphic} to improve the generation performance. In the early stage, traditional models \cite{lian2012automatic,liu2012automatic,lin2015complete} firstly decompose the Chinese characters into several basic components such as strokes and radicals, and then assemble them to yield new Chinese characters by leveraging some effective machine learning methods. It should be pointed out that the decomposition of strokes or radicals is generally hand-crafted and thus very costly.

To address this issue, several end-to-end models based on deep learning are introduced to extract these important components of Chinese characters and incorporate them into the generation procedure \cite{lin2019font,wen2021handwritten,https://doi.org/10.48550/arxiv.2204.10484,jiang2019scfont,gao2020gan,qin2022disentangled,zeng2021strokegan,zeng2022SQ-GAN}. Specifically, \cite{lin2019font} firstly adopted a coherent point drift algorithm to divide the Chinese characters into strokes, and then produced new font strokes by fusing the styles of two existing font strokes, and further yielded new fonts through assembling them. \cite{wen2021handwritten} suggested a stroke refinement branch to particularly focus on the generation of thin strokes. \cite{zeng2021strokegan} introduced a simple one-bit stroke encoding to represent the stroke information and incorporated it into the CycleGAN model as the supervision information. By the use of such stroke encoding, the mode collapse of CycleGAN can be remarkably reduced \cite{zeng2021strokegan}. In the recent paper \cite{zeng2022SQ-GAN}, the authors proposed a square-block translation based self-supervised scheme for the generation of Chinese characters, inspired by the fact that structures of Chinese characters are closely related to the square-block transformations. As compared to \cite{lin2019font,wen2021handwritten,zeng2021strokegan,zeng2022SQ-GAN}, this paper exploits the skeleton instead of the stroke representing certain local structure information \cite{lin2019font,wen2021handwritten,zeng2021strokegan} and the square-block transformation representing certain global structure information \cite{zeng2022SQ-GAN}, mainly motivated by the observation that skeletons can provide both local and global structure information of Chinese characters. 

Besides the stroke and character structure information, several works based on the skeleton-like information have been studied in the literature \cite{jiang2019scfont,gao2020gan,Y2021Calligraphic,https://doi.org/10.48550/arxiv.2204.10484,qin2022disentangled}. \cite{jiang2019scfont} introduced two neural network modules to realize the extraction of strokes and writing trajectories of Chinese characters respectively, and then incorporated them into the generation of Chinese fonts. \cite{gao2020gan} proposed a three-stage GAN model including the skeleton extraction, skeleton transformation and stroke rendering for multi-chirography image translation. \cite{qin2022disentangled} proposed a font fusion network based on GAN and disentangled representation learning to generate brand new fonts, where the disentangled representation learning was introduced to obtain the stoke style and skeleton shape. \cite{Y2021Calligraphic} incorporated the skeleton of Chinese characters into the generator and utilized it as certain structural constraint to supervise the model, while similar idea was also used in \cite{https://doi.org/10.48550/arxiv.2204.10484} for the generation of brush handwriting Chinese font. It should be pointed out that both \cite{Y2021Calligraphic} and \cite{https://doi.org/10.48550/arxiv.2204.10484} are based on the paired data and through introducing auxiliary network modules to extract skeletons, while this paper is based on the unpaired data and through using some simple mathematical verification conditions to extract the skeleton. 

\begin{table}[h!]
\large
\centering
\begin{tabular}{|c|c|c|} 
 \hline
$P_9$ & $P_2$ & $P_3$ \\
 \hline
 $P_8$ & $P_1$ & $P_4$ \\
 \hline
 $P_7$ & $P_6$ & $P_5$ \\
 \hline
\end{tabular}
\caption{Layout of the defined 3 $\times$ 3 image patch.}
\label{table:window}
\end{table}

Distinguished from previous studies struggling to increase the model complexity to utilize these components of Chinese characters to help the generation tasks, we aim to perform much simpler and more effective supervision for a relatively simple GAN model to achieve its full potential. As discussed before, the proposed skeleton guided channel expansion module is very simple but effective to provide certain local and global supervision for GAN models to alleviate the mode collapse issue and enhance the performance, while does not introduce any additional model complexity. 
The skeleton information used in this paper can provide richer spatial structure information and semantic information of Chinese characters, and the suggested channel expansion way to exploit the skeleton information can provide stronger guidance for the generator.
Moreover, the proposed module can be easily adapted to many existing GAN models to improve their performance.

\section{SGCE for Chinese Font Generation}
\label{sc:proposed model}
In this section, we firstly describe the proposed skeleton guided channel expansion (SGCE) module as an effective information guidance module for the Chinese font generation, and then take the well-known CycleGAN \cite{chang2018generating} as an example to show how to integrate the proposed module into the existing Chinese font generation models.


\subsection{Skeleton Guided Channel Expansion Module}
\label{sc:SGCE}
As discussed before, using what kind of information as the guidance information and how to use it are very important for the Chinese font generation. Motivated by the superiority of the skeleton on representing the global and local structure information of Chinese characters and the advantage of channel expansion on directly using the information, we propose a skeleton guided channel expansion module as certain a information guidance module for the Chinese font generation, as depicted in Figure \ref{fig:SGCE}. From Figure \ref{fig:SGCE}, given a character image $x$ as the input, we firstly extract its skeleton $sx$ with single channel via a skeletonization strategy denoted as \textit{Ske} for short and described later, and then combine the single-channel skeleton information with the original (R, G, B) three-channel character image information to yield the four-channel image information $cex$ as the input of the generator through the channel expansion way.

In the next, we describe the skeletonization strategy used in this paper to extract the skeleton of a Chinese character, inspired by \cite{zhang1984fast}.
Given an image of a Chinese character, we firstly perform a binarized operation on it, and then yield the skeleton from the binarized image. Specifically, for a given pixel $P_1$ of the binarized image, we define a $3\times 3$ image patch with $P_1$ as the center, and number the other eight pixels according to Table \ref{table:window}. For those pixels lying on the edge of the binarized image, we obtain the associated $3\times 3$ image patches using the zero-padding way. For the defined $3\times 3$ image patch as shown in Table \ref{table:window}, let $p_i$ be the value of $P_i$ ($i=1,\ldots,9$), ${\cal N}(P_1)$ be the number of non-zero points in this $3\times 3$ image patch, and ${\cal P}(P_1)$ be the number of $(0,1)$ pairs among $P_2$ to $P_9$ by sequence, then we delete $P_1$ if the following hold:
\begin{enumerate}
    \item[(a)] $2 \leq {\cal N}(P_1)\leq 6, \  {\cal P}(P_1)=1$; and either
    \item[(b)] $p_2 \cdot p_4 \cdot p_6=0, \ p_4 \cdot p_6 \cdot p_8=0$; or
    \item[(c)] $p_2 \cdot p_4 \cdot p_8=0, \ p_2 \cdot p_6 \cdot p_8=0$.
\end{enumerate}
We eventually yield the skeleton of a Chinese character through doing the above operation over all pixels.

According to the above description, the proposed SGCE module does not require any additional neural network to implement it. This is very different to existing works \cite{jiang2019scfont,gao2020gan,Y2021Calligraphic,https://doi.org/10.48550/arxiv.2204.10484,qin2022disentangled} struggling to increase the model complexity to exploit the component information (say skeleton) for the Chinese font generation.

\subsection{Proposed SGCE-Font Model}
\label{sc:architecture}
In the following, we describe the proposed model for Chinese font generation through integrating the suggested SGCE module (called \textit{SGCE-Font}), where we take the well-known CycleGAN \cite{chang2018generating} as the base model as an example. The specific model architecture is presented in Figure \ref{fig:SGCE-Font}. As depicted in Figure \ref{fig:SGCE-Font}, the proposed SGCE-Font is very similar to that of the existing model (say, CycleGAN), except that the input of the generator is replaced by the four-channel skeleton-image information yielded by the suggested SGCE module, instead of the original (R, G, B) three-channel image information.
The specific workflow of the proposed model can be described as follows. We first input the Chinese character image $x$ with (R, G, B) three channels in the source font domain into SGCE module to yield $cex = SGCE(x)$ enriched by the skeleton information into the generator $G_y$ to realize the style translation from the source font style to the target font style. 
After $G_y$, we yield the generated character $\tilde{y} = G_y(cex)$ in the target font domain. On one hand, we simultaneously feed the generated characters and the real Chinese characters in the target font domain into the discriminator $D_y$ to distinguish whether they are real or fake, and on the other hand, we feed the generated character $\tilde{y} = G_y(cex)$ into the SGCE module and then input the enriched information into another generator $G_x$ (realizing the font generation task from the target font domain to the source font domain), and finally yield a reconstructed character $\tilde{x} = G_x(SGCE(\tilde{y}))$ in the source font domain. 
According to the above workflow, the training loss of the proposed model consists of four parts including two adversarial losses ${\cal L}_{adv_x}$ and ${\cal L}_{adv_y}$ related to $(D_x, G_x)$ and $(D_y, G_y)$ respectively, the cycle consistency loss ${\cal L}_{cyc}$ between the original Chinese character $x$ and its reconstructed character $\tilde{x}$ in the source font domain, the skeleton consistency loss ${\cal L}_{ske}$ between the skeleton $sx$ of the input $x$ and the skeleton $\widetilde{sx} = Ske(\tilde{x})$ of the reconstructed Chinese character $\tilde{x}$ in the source font domain, where the skeleton consistency loss is imposed to further supervise the generation procedure. Specifically, these losses are defined as follows:
\begin{align*}
&\mathcal{L}_{adv_{x}}\left(D_{x}, G_{x}\right) \\
&=E_{x \sim \mathcal{X}}\left[\log D_{x}(x)\right]+E_{\tilde{x} \sim \tilde{\mathcal{X}}}\left[\log \left(1-D_{x}(\tilde{x})\right)\right], \\
&\mathcal{L}_{a d v_{y}}\left(D_{y}, G_{y}\right) \\
&=E_{y \sim \mathcal{Y}}\left[\log D_{y}(y)\right]+E_{\tilde{y} \sim \tilde{\mathcal{Y}}}\left[\log \left(1-D_{y}(\tilde{y})\right)\right], 
\end{align*}
\begin{align*}
&\mathcal{L}_{cyc}\left(G_{x}, G_{y}\right) =E_{(x,\tilde{x}) \sim(\mathcal{X}, \tilde{\mathcal{X}})}\|x-\tilde{x}\|_{1},\\
&\mathcal{L}_{ske}\left(G_{x}, G_{y}\right) 
=E_{(x, \tilde{x}) \sim(\mathcal{X}, \tilde{\mathcal{X}})}\|Ske(x)-{Ske}(\tilde{x})\|_{1},
\end{align*}
where ${\cal X}$ and ${\cal Y}$ respectively represent the sets of real Chinese characters in the source and target domains, $\tilde{\cal X}$ represents the set of reconstructed Chinese characters in the source font domain by $G_x$, $\tilde{\cal Y}$ represents the set of generated characters in the target font domain by $G_y$. Thus, the training model of SGCE-Font is shown as follows:
\begin{align*}
    \min_{G:=(G_x,G_y)} \max_{D:=(D_x,D_y)} {\cal L}_{SGCE-Font}(D,G),
\end{align*}
where 
\begin{align*}
 &{\cal L}_{SGCE-Font}(D,G) 
 = {\cal L}_{adv_x}(D_x, G_x) +  {\cal L}_{adv_y}(D_y, G_y) \\
 &+ \lambda_{cyc}{\cal L}_{cyc}(G_x,G_y) + \lambda_{ske}  {\cal L}_{ske}(G_x,G_y), 
\end{align*}
and $\lambda_{cyc}$, $\lambda_{ske}$ are two tunable hyperparameters.
\begin{table*}[!h]
\begin{center}
\begin{tabular}{|c|c|c|c|c|c|c|c|c|c|c|}\hline
 Font  & ht   & fs    & zk   & hw  & hp  & st  & ls  & hc   & zx    & ss  \\ \hline
 Size  &3755  &3755   &2811  &3755 &2811 &2811 &2811 &2811  & 920   & 166 \\ \hline 
  \end{tabular}
  \caption{Number of sizes of different font datasets.}
    \label{table:fontsize}
\end{center}
\end{table*}

\begin{table*}[!htbp]
\small
\centering
\setlength{\tabcolsep}{0.29mm}{
\begin{tabular}{c|c|c|c|c|c|c|c|c|c|c|c|c|c|c|c|c}
\hline
& task & zk$\rightarrow$fs & zk$\rightarrow$st  & fs$\rightarrow$ht  & zk$\rightarrow$hp & zk$\rightarrow$hw & fs$\rightarrow$ls & ls$\rightarrow$hc  & fs$\rightarrow$zk & st$\rightarrow$zk  & ht$\rightarrow$fs  & hp$\rightarrow$zk  & hw$\rightarrow$zk & ls$\rightarrow$fs & hc$\rightarrow$ls   & avg     \\
\hline
\multirow{5}{*}{\rotatebox{90}{FID}}   & CycleGAN  & \textbf{25.80} & 79.33 & 304.34 & 43.91 & \textcolor{blue}{31.54} & 282.81 & \textcolor{blue}{33.02}  & \textcolor{blue}{32.56} &47.19  & 248.63 & 218.18 & 27.10 & 136.13 & \textcolor{blue}{30.17} &110.05 \\
& SQ-GAN    & 49.87 & 71.69 & 38.90  & 57.44 & 49.53 & 29.08 &71.22  & 99.04 & 146.29 & \textcolor{blue}{37.55}  & 305.93 & 24.24 & 56.90  & 50.45 &77.72  \\
& StrokeGAN & 57.85 & 71.30 & \textcolor{blue}{28.97}  & 92.86 & 37.78 & \textcolor{blue}{28.63} &34.60  & 67.51 & 71.30  & 39.61  & 140.23 & \textcolor{blue}{23.80} & \textbf{27.35}  & \textbf{26.74}  &53.47 \\
& SkeGAN    & 36.58 & \textcolor{blue}{58.75} & 40.27  & \textbf{26.90} &\textbf{28.91} & 32.32 &46.09 & 33.58 & \textcolor{blue}{33.12} & 59.52  & \textcolor{blue}{27.59}  & 48.29 & 49.84  & 72.97  &\textcolor{blue}{42.48} \\
& SGCE-Font      & \textcolor{blue}{34.68} & \textbf{40.35} & \textbf{20.66}  & \textcolor{blue}{34.60} & 53.57 & \textbf{20.79} &\textbf{30.29} & \textbf{25.55} & \textbf{31.91}  & \textbf{23.73}  & \textbf{26.74}  & \textbf{23.03} & \textcolor{blue}{35.70}  & 30.70 &\textbf{30.88}\\
\hline\hline
\multirow{5}{*}{\rotatebox{90}{MSE}}  & CycleGAN  & 0.117  & 0.214  & 0.197   & 0.212  & 0.097  & 0.187  &\textcolor{blue}{0.155} & 0.135  & \textbf{0.121}   & 0.166   & 0.177   & 0.201  & 0.193   & 0.148  &0.166  \\
& SQ-GAN    & 0.147  & 0.199  & 0.186   & 0.292  & 0.106  & 0.162 &0.176  & 0.152  & 0.158   & 0.143   & 0.161   & 0.208  & 0.153   & 0.206 &0.175   \\
& StrokeGAN & 0.120  & 0.158  & 0.188   & 0.223  & 0.104  & \textcolor{blue}{0.090}  &0.175 & \textcolor{blue}{0.104}  & 0.125   & 0.129   & 0.181   & 0.207  & \textcolor{blue}{0.116}   & \textbf{0.112}  &0.145  \\
& SkeGAN    & \textcolor{blue}{0.112}  & \textcolor{blue}{0.198}  & \textbf{0.106}   & \textcolor{blue}{0.187}  & \textcolor{blue}{0.092}  & \textbf{0.088} &\textbf{0.150}  & 0.133  & 0.128   & \textbf{0.116}   & \textcolor{blue}{0.121}   & \textcolor{blue}{0.191}  & \textbf{0.111}   & 0.146  &\textcolor{blue}{0.134}  \\
& SGCE-Font      & \textbf{0.111}  & \textbf{0.158}  & \textcolor{blue}{0.117}   & \textbf{0.175}  & \textbf{0.090}  & 0.094  &0.163 & \textbf{0.087}  & \textcolor{blue}{0.122}   & \textcolor{blue}{0.117}   & \textbf{0.112}   & \textbf{0.188}  & 0.123   & \textcolor{blue}{0.144}   &\textbf{0.129} \\
\hline\hline
\multirow{5}{*}{\rotatebox{90}{PSNR}}  & CycleGAN  & 9.51  & \textcolor{blue}{7.26}  & 7.07   & 6.79  & 10.21 & 7.287 &\textcolor{blue}{8.16}  & 8.75  & 8.32   & 7.80   & 7.55   & 7.02  & 7.17   & 5.65   &7.75 \\
& SQ-GAN    & \textbf{9.82}  & 7.06  & 7.82   & 6.98  & 9.79  & 9.05 &7.72  & 9.37  & 8.05   & 8.63   & 7.40   & 6.85  & 8.99   & 8.36  &8.28  \\
& StrokeGAN & 9.36  & 6.77  & 7.32   & 6.35  & 9.91  & \textcolor{blue}{10.66} &7.72 & \textcolor{blue}{10.00} & 6.77   & 9.04   & 8.38   & 6.88  & \textcolor{blue}{9.55}   & \textbf{9.68}  &8.46  \\
& SkeGAN    & 9.70  & 7.08  & \textbf{9.98}   & \textcolor{blue}{7.37}  & \textcolor{blue}{10.46} & \textbf{10.81} &\textbf{8.30} & 8.83  & \textcolor{blue}{9.06}   & \textcolor{blue}{9.47}   & \textcolor{blue}{9.29}   & \textcolor{blue}{7.25}  & \textbf{9.73}   & 8.42 &\textcolor{blue}{8.98}  \\
& SGCE-Font      & \textcolor{blue}{9.73}  & \textbf{8.11}  & \textcolor{blue}{9.55}   &\textbf{7.72}  & \textbf{10.55} & 10.47 &7.95 & \textbf{10.85} & \textbf{9.32}   & \textbf{9.69}   & \textbf{9.68}   & \textbf{7.35}  & 9.28   & \textcolor{blue}{8.49}  &\textbf{9.20} \\
\hline\hline
\multirow{5}{*}{\rotatebox{90}{SSIM}}  & CycleGAN  & 0.568  & \textcolor{blue}{0.495}  & 0.346   & 0.386  & 0.330  & 0.461 &0.519  & 0.502  & 0.473   & 0.146   & 0.413   & 0.459  & 0.433   & 0.008 &0.396\\
& SQ-GAN    & 0.570  & 0.448  & 0.478   & 0.354  & 0.293  & 0.595 &0.499  & 0.530  & 0.428   & 0.485   & 0.359   & 0.444  & 0.505   & 0.543  &0.466  \\
& StrokeGAN & 0.553  & 0.464  & 0.462   & 0.300  & 0.301  & \textcolor{blue}{0.660}  &0.452 & \textcolor{blue}{0.574}  & 0.464   & \textcolor{blue}{0.533}   & 0.448   & 0.447  & \textcolor{blue}{0.547}   & \textcolor{blue}{0.566}  &0.484  \\
& SkeGAN    & \textcolor{blue}{0.579}  & 0.465  & \textcolor{blue}{0.605}   & \textcolor{blue}{0.428}  & \textcolor{blue}{0.341}  & 0.657 &\textcolor{blue}{0.535}  & 0.506  & \textcolor{blue}{0.527}   & 0.526   & \textcolor{blue}{0.543}   & \textcolor{blue}{0.460}  & 0.538   & 0.553 &\textcolor{blue}{0.519}  \\
& SGCE-Font      & \textbf{0.582}  & \textbf{0.534}  & \textbf{0.613}   & \textbf{0.452}  & \textbf{0.349}  & \textbf{0.664} &\textbf{0.539}  & \textbf{0.632}  & \textbf{0.542}   & \textbf{0.579}   &\textbf{ 0.566}   & \textbf{0.476}  & \textbf{0.549}   & \textbf{0.578} &\textbf{0.547}  \\
\hline
\end{tabular}
\caption{Comparison on the performance of the proposed SGCE-Font and existing GAN models with other guidance schemes over fourteen font generation tasks. The best and second best results are marked in bold and blue color respectively.}
\label{table:comparisonasfont_ofSK_SR_SQ}
}
\end{table*}

\section{Experiments}
\label{sc:experiments}

In this section, we conducted a series of experiments to show the effectiveness of the proposed SGCE module for the Chinese font generation. All experiments were carried out in Pytorch environment running Linux, AMD Ryzen 7 5800X 8-Core Processor CPU, GeForce RTX 3090 GPU.

\begin{figure*}[!h]
\begin{minipage}[b]{0.99\linewidth}
\centering
\includegraphics*[scale=0.22]{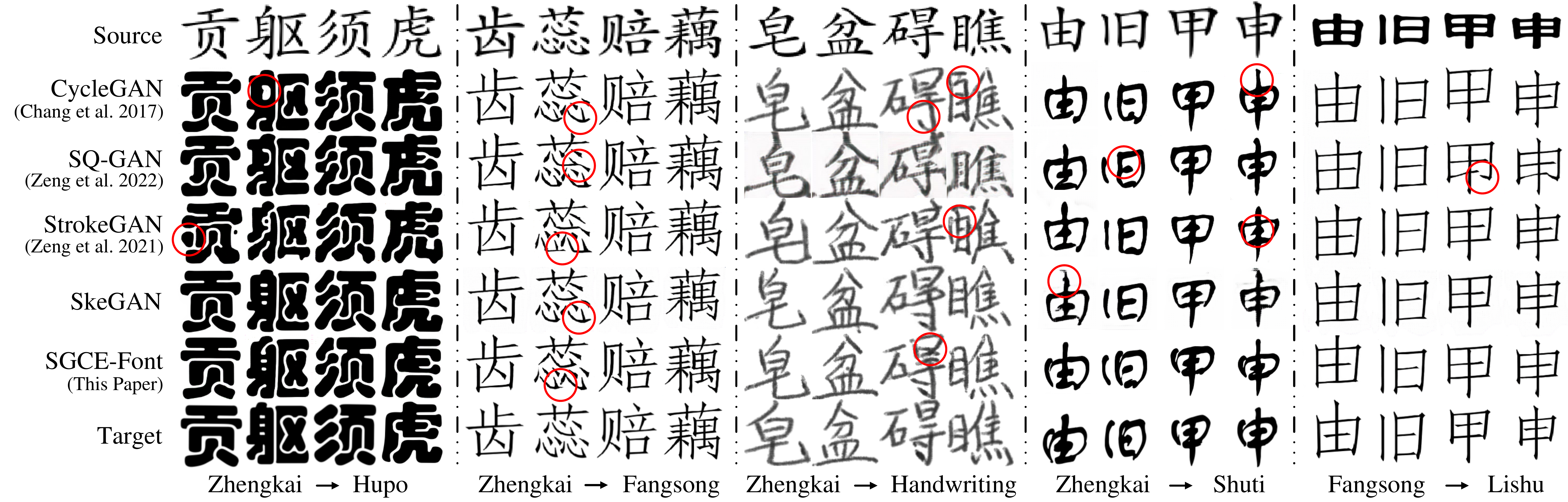}
\end{minipage}
\hfill
\caption{Generated characters of SGCE-Font and baselines CycleGAN \cite{zhu2017unpaired}, SQ-GAN \cite{zeng2022SQ-GAN}, StrokeGAN \cite{zeng2021strokegan} and SkeGAN over five generation tasks.}
\label{fig:SkeGAN-CycleGAN-SQ-GAN-StrokeGAN}
\end{figure*}

\subsection{Experimental Settings}
\label{sc:experiment settings}

\textbf{A. Datasets.}
We considered ten datasets with different fonts including three standard printing fonts \{Heiti (ht), Fangsong (fs), Zhengkai (zk)\}, the handwriting (hw) font, three pseudo-handwriting fonts \{Hupo (hp), Shuti (st), Lishu (ls)\}, and three calligraphy fonts \{Huangcao (hc), Zhuxi (zx), SuShi(ss)\}. The specific sizes of datasets were presented in Table \ref{table:fontsize} and some samples were shown in the left part of Figure \ref{fig:generated-characters}. The handwriting dataset was randomly collected from the CASIA-HWDB1.1 dataset, the other font datasets were collected by ourselves from the internet. The sizes of the last two calligraphy font datasets, i.e., Zhuxi and SuShi are very small. We use them mainly to show the performance of the proposed model over the few-shot font generation task. For each sample, we reset the image size of each character to 128×128×3. In all experiments, we used 80\% and 20\%  samples as the training and test sets respectively.

\textbf{B. Baselines.} In this paper, we considered the following six state-of-the-art models as baselines.
\begin{itemize}
\item \textbf{CycleGAN \cite{chang2018generating}:} A typical model based on GAN and unpaired data.

\item \textbf{SQ-GAN \cite{zeng2022SQ-GAN}:} An improved CycleGAN model equipped with a square-block geometric transformation based self-supervised learning scheme.

\item \textbf{StrokeGAN \cite{zeng2021strokegan}:} A refined CycleGAN model incorporated with the one-bit stroke encoding for the Chinese font generation to mitigate the mode collapse.

\item \textbf{UGATIT \cite{isola2017image}:} An effective GAN model using unsupervised generative attention network with adaptive layer instance normalization and unpaired data.

\item \textbf{FUNIT \cite{liu2019few}:} An effective GAN model based on the disentangled representation learning and unpaired data.

\item \textbf{AttentionGAN \cite{tang2021attentiongan}:} An effective GAN model based on the attention mechanism and unpaired data.
\end{itemize}

Besides the above baselines, we also suggested another model called \textbf{SkeGAN} as a baseline for better comparison with the proposed model. In the SkeGAN, we impose the skeleton information to the discriminator in the similar way of StrokeGAN \cite{zeng2021strokegan}.

\textbf{C. Network structures.}
The network structure of the generator of SGCE-Font is almost the same as that of CycleGAN \cite{zhu2017unpaired}, including 2 convolutional layers in the down-sampling module, 9 residual modules with 2 convolutional layers of residual networks for each residual module, and 2 deconvolutional layers in the up-sampling module, except that the input of the generator was modified from 3 channels to 4 channels. The network structure of the discriminator of SGCE-Font is the same as CycleGAN \cite{zhu2017unpaired}, with 6 hidden convolutional layers and 2 convolutional layers in the output module. Moreover, batch normalization \cite{ioffe2015batch} was used in all layers.

We used the popular Adam algorithm \cite{kingma2014adam} as the optimizer with the associated parameters (0.5, 0.999) in both the generator and discriminator optimization subproblems. The hyperparameters of the cycle consistency loss and skeleton consistency loss were empirically set as 1 and 0.001, respectively.

\textbf{D. Evaluation metrics.}
We used four important evaluation metrics to measure the performance. The first two evaluation metrics were the commonly used SSIM (Structural Similarity) \cite{Wang2004-SSIM} and PSNR (Peak Signal to Noise Ratio), which were employed to measure how the pixel-level details are preserved. The third evaluation metric is the well-known MSE (Mean Square Error). The last evaluation metric is FID (Frechet Inception Distance) \cite{heusel2017gans}, which was used to measure how the generated results of the model match the distribution of real data. Larger SSIM and PSNR values, and smaller MSE and FID values generally imply better generation performance. 
\begin{table*}[!htbp]
\centering
\small
\setlength{\tabcolsep}{0.9mm}{
\begin{tabular}{c|c|c|c|c|c|c|c|c|c|c|c|c|c|c}
\hline
& task     & zk$\rightarrow$fs & zk$\rightarrow$st  & fs$\rightarrow$ht  & zk$\rightarrow$hp & zk$\rightarrow$hw & fs$\rightarrow$ls   & fs$\rightarrow$zk & st$\rightarrow$zk  & ht$\rightarrow$fs  & hp$\rightarrow$zk  & hw$\rightarrow$zk & ls$\rightarrow$fs  & avg     \\
\hline
\multirow{4}{*}{\rotatebox{90}{FID}}   & FUNIT        & 37.50 & \color{blue}43.67  & 64.99 & 55.28 & 75.55 & 30.84  & 151.73 & 207.22 & 40.68 & 254.40 & 68.82 & 63.33  & 91.17 \\
& UGATIT       & 42.82 & 118.10 & 37.92 & 64.37 & 77.10 & 30.32  &\color{blue} 34.03  & 128.10 &\color{blue} 28.32 & 76.55  & 29.01 & \textbf{23.00}  & 58.11  \\
& AttentionGAN &\textbf{32.32} & 92.03  &\color{blue} 32.80 & \textbf{25.20} & \textbf{26.86} &\textbf{17.77} & 57.46  &\color{blue} 51.84  & 37.74 & \color{blue}55.69  &\color{blue} 25.01 & \color{blue}34.33  & \textcolor{blue}{40.76}  \\
& SGCE-Font        &\color{blue} 34.68 & \textbf{40.35}  & \textbf{20.66} & \color{blue}34.60 &\color{blue} 53.57 &\color{blue} 20.79  &\textbf{25.55}  &\textbf{31.91}  & \textbf{23.73} &\textbf{26.73}  & \textbf{23.03} & 35.70 &\textbf{30.94}\\
\hline\hline
\multirow{4}{*}{\rotatebox{90}{MSE}}  & FUNIT        & 0.127  & 0.224   & 0.212  & 0.331  & 0.105  & 0.170    & 0.138   & 0.180   & 0.178  & 0.188   & 0.216  &0.155  &0.185  \\
& UGATIT       & 0.113  & \textbf{0.151}   &\color{blue} 0.170  & 0.263  & 0.129  & 0.119    &\color{blue} 0.090   &\color{blue} 0.127   & 0.140  & 0.144   & 0.224  & 0.133   &0.150   \\
& AttentionGAN & \textbf{0.109}  & 0.197   &\color{blue} 0.170  & \textbf{0.162}  &\color{blue} 0.097  & \textbf{0.087}   & 0.095   & 0.151   &\color{blue} 0.125  &\color{blue} 0.143   &\color{blue} 0.191  &\color{blue} 0.126  &\textcolor{blue}{0.138}   \\
& SGCE-Font         &\color{blue} 0.111  & \color{blue}0.158   & \textbf{0.117}  & \color{blue}0.175  & \textbf{0.090}  &\color{blue} 0.094   & \textbf{0.087}   & \textbf{0.122}   & \textbf{0.117}  & \textbf{0.112}   & \textbf{0.188}  & \textbf{0.123}  &\textbf{0.125}   \\
\hline\hline
\multirow{4}{*}{\rotatebox{90}{PSNR}}  & FUNIT        & 9.07  & 6.52   & 6.77  & 4.83  & 9.86  & 7.75  & 8.66   & 7.48   & 7.52  & 7.28   & 6.68  & 8.16    &7.56  \\
& UGATIT       & 9.63  & \textbf{8.34}   &\color{blue} 7.80  & 5.87  & 9.34  & 8.83   &\color{blue} 10.65  &\color{blue} 9.10   & 8.66  &\color{blue} 8.48   & 6.55  & \textbf{9.43} &8.52   \\
& AttentionGAN & \textbf{9.83}  & 7.08   & 7.75  &\textbf{8.05} & \color{blue}10.24 &\textbf{10.86}   & 10.40  & 8.27   &\color{blue} 9.15  & 8.47   & \color{blue}7.26  & 9.09   &\textcolor{blue}{8.87}   \\
& SGCE-Font          &\color{blue} 9.73  &\color{blue} 8.11   & \textbf{9.55}  & \color{blue}7.71  & \textbf{10.55}& \color{blue}10.47   & \textbf{10.85}  &\textbf{9.32}   & \textbf{9.69}  & \textbf{9.68}   & \textbf{7.35} & \color{blue}9.28   & \textbf{9.36}   \\
\hline\hline
\multirow{4}{*}{\rotatebox{90}{SSIM}}  & FUNIT        & 0.525 & 0.426  & 0.388 & 0.241 & 0.295 & 0.511 & 0.485  & 0.412  & 0.389 & 0.386  & 0.373 & 0.460 & 0.408  \\
& UGATIT       & 0.549 & \textcolor{blue}{0.528}  & \textcolor{blue}{0.549} & 0.341 & 0.271 & 0.642 &\textcolor{blue}{0.589}  & \textcolor{blue}{0.486}  & \textcolor{blue}{0.549} & 0.451  & 0.400 & \textbf{0.607} & \textcolor{blue}{0.497}  \\
& AttentionGAN & \textcolor{blue}{0.580} & 0.451  & 0.463 & \textcolor{blue}{0.440} & \textcolor{blue}{0.332} & \textbf{0.665} & 0.588  & 0.433  & 0.507 & \textcolor{blue}{0.452}  & \textcolor{blue}{0.462} & 0.505 & 0.490  \\
& SGCE-Font         & \textbf{0.582} & \textbf{0.534}  & \textbf{0.613} & \textbf{0.452} & \textbf{0.349} & \textcolor{blue}{0.664} & \textbf{0.632}  & \textbf{0.542}  & \textbf{0.579} & \textbf{0.566}  & \textbf{0.476} & \textcolor{blue}{0.549} & \textbf{0.545}   \\
\hline
\end{tabular}
\caption{Comparison on the performance of SGCE-Font and state-of-the-art models over fourteen generation tasks. The best and second best results are marked in bold and blue color respectively.}
\label{table:comparison-sota}
}
\end{table*}

\subsection{Superiority of Proposed SGCE Module}
\label{sc:skeleton vs stroke}
In this section, we implemented extensive experiments over fourteen generation tasks to demonstrate the superiority of the proposed SGCE module for the Chinese font generation. The quantitative comparisons between the proposed SGCE-Font and these baselines including CycleGAN \cite{chang2018generating}, SQ-GAN \cite{zeng2022SQ-GAN}, StrokeGAN \cite{zeng2021strokegan} and SkeGAN are presented in Table \ref{table:comparisonasfont_ofSK_SR_SQ}, and some visual comparisons on the generated characters are presented in Figure \ref{fig:modecollapse} and Figure \ref{fig:SkeGAN-CycleGAN-SQ-GAN-StrokeGAN}.

From Table \ref{table:comparisonasfont_ofSK_SR_SQ}, the proposed SGCE-Font model outperforms these concerned baselines in terms of four evaluation metrics in average. Specifically, when regarding the FID, the proposed SGCE-Font achieves the best results in nine font generation tasks and the second best results in three font generation tasks, and is significantly better than the baselines. Similar claims can be concluded in terms of the other three evaluation metrics. 

When comparing the performance among different models, the refined CycleGAN models using different guidance information such as the square-block transformation in SQ-GAN \cite{zeng2022SQ-GAN}, the stroke encoding in StrokeGAN \cite{zeng2021strokegan} and the skeleton in SkeGAN outperform the original CycleGAN model for the Chinese font generation.
Considering the performance of these models using different guidance information, the performance of SkeGAN using the skeleton information is significantly better than that of SQ-GAN and StrokeGAN in average, in terms of these four evaluation metrics. This show clearly that the skeleton information is much better than the square-block transformation and stroke encoding as the guidance information for the Chinese font generation.
Moreover, when comparing the performance of SkeGAN and the proposed SGCE-Font, it can be observed that the proposed SGCE-Font outperforms SkeGAN in most of cases. This shows that the suggested way of using the skeleton information by the channel expansion as the direct input of the generator is much better than the way of imposing the skeleton information onto the discriminator as used in SkeGAN. 
These claims can be also verified by the visualization results as depicted in Figure \ref{fig:SkeGAN-CycleGAN-SQ-GAN-StrokeGAN}.
From Figure \ref{fig:SkeGAN-CycleGAN-SQ-GAN-StrokeGAN}, the quality of generated characters of the proposed SGCE-Font is better than SkeGAN, while the performance of SkeGAN is much better than that of StrokeGAN, SQ-GAN as well as CycleGAN.

In particular, as shown in Figure \ref{fig:modecollapse}, the well-known CycleGAN suffers from the mode collapse issue when applied to these three font generation tasks, i.e., \{Hupo $\rightarrow$ Zhengkai, Fangsong $\rightarrow$ Heiti, Fangsong $\rightarrow$ Lishu\}. Such mode collapse phenomena of CycleGAN can be also observed from Table \ref{table:comparisonasfont_ofSK_SR_SQ}, where the FID values of CycleGAN in these three font generation tasks are 218.18, 304.34 and 282.81, which are abnormally large. Although existing square-block transformation and stroke encoding based guidance schemes can reduce the mode collapse to some extent as shown by the third and fourth rows of Figure \ref{fig:modecollapse}, the SQ-GAN \cite{zeng2022SQ-GAN} still suffers from the mode collapse in the font generation task \{Hupo $\rightarrow$ Zhengkai\} and there is stroke-missing or stroke-redundancy phenomenon in the generated characters of StrokeGAN,
while from the fifth rows of Figure \ref{fig:modecollapse}, the mode collapse issue of CycelGAN can be significantly alleviated by the suggested SGCE module, and the quality of the generated characters of the proposed SGCE-Font is much better than that of the other three baselines. These show clearly the superiority of the suggested SGCE module on reducing mode collapse. 

\begin{figure*}[!t]
\begin{minipage}[b]{0.95\linewidth}
\centering
\includegraphics*[scale=0.179]{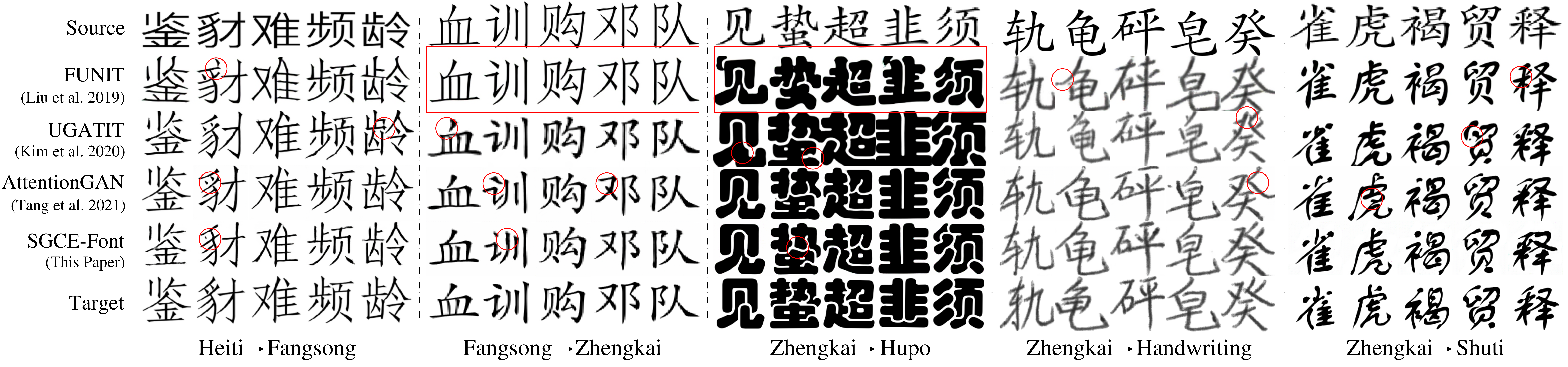}
\end{minipage}
\hfill
\caption{Comparison on the generated quality of the proposed SGCE-Font model and three state-of-the-art models FUNIT \cite{liu2019few}, UGATIT \cite{Kim2020UGATITUG} and AttentionGAN \cite{tang2021attentiongan} over five font generation tasks and for the commonly used characters.
}
\label{fig:comparison-sota}
\end{figure*}

\begin{figure*}[htbp]
\begin{minipage}[b]{0.99\linewidth}
\centering
\includegraphics*[scale=0.25]{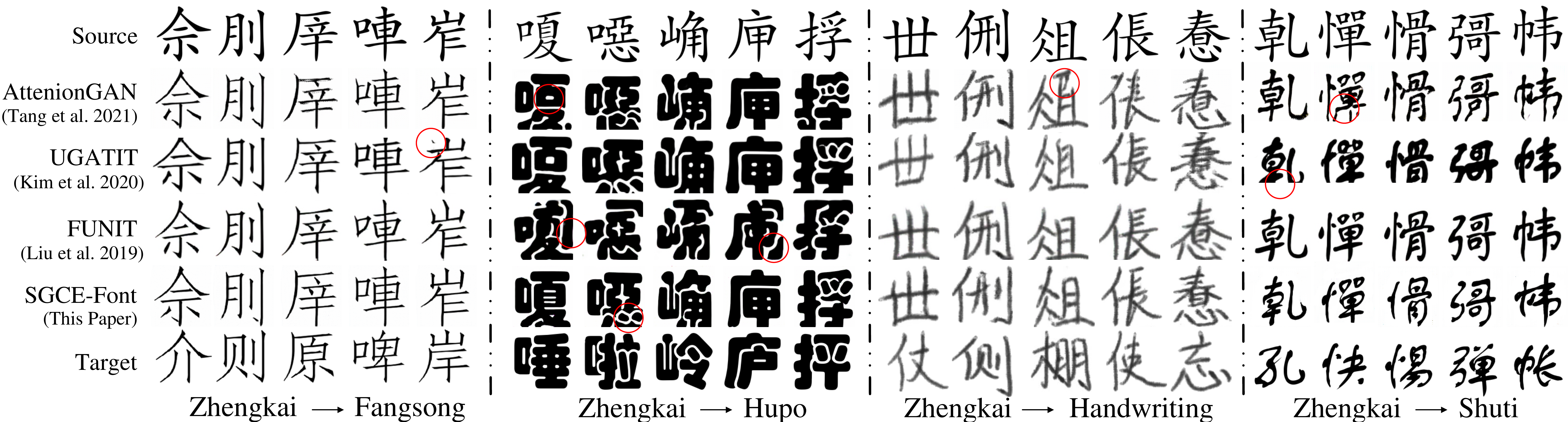}
\end{minipage}  
\hfill
\caption{Comparison on the generalization performance of the proposed SGCE-Font model and three state-of-the-art models FUNIT \cite{liu2019few}, UGATIT \cite{Kim2020UGATITUG} and AttentionGAN \cite{tang2021attentiongan} over four font generation tasks and for the unseen characters (i.e., excluded by the concerned datasets). }
\label{fig:comparison-sota-unseenfont}
\end{figure*}

\begin{figure*}[htbp]
\begin{minipage}[b]{0.99\linewidth}
\centering
\includegraphics*[scale=0.25]{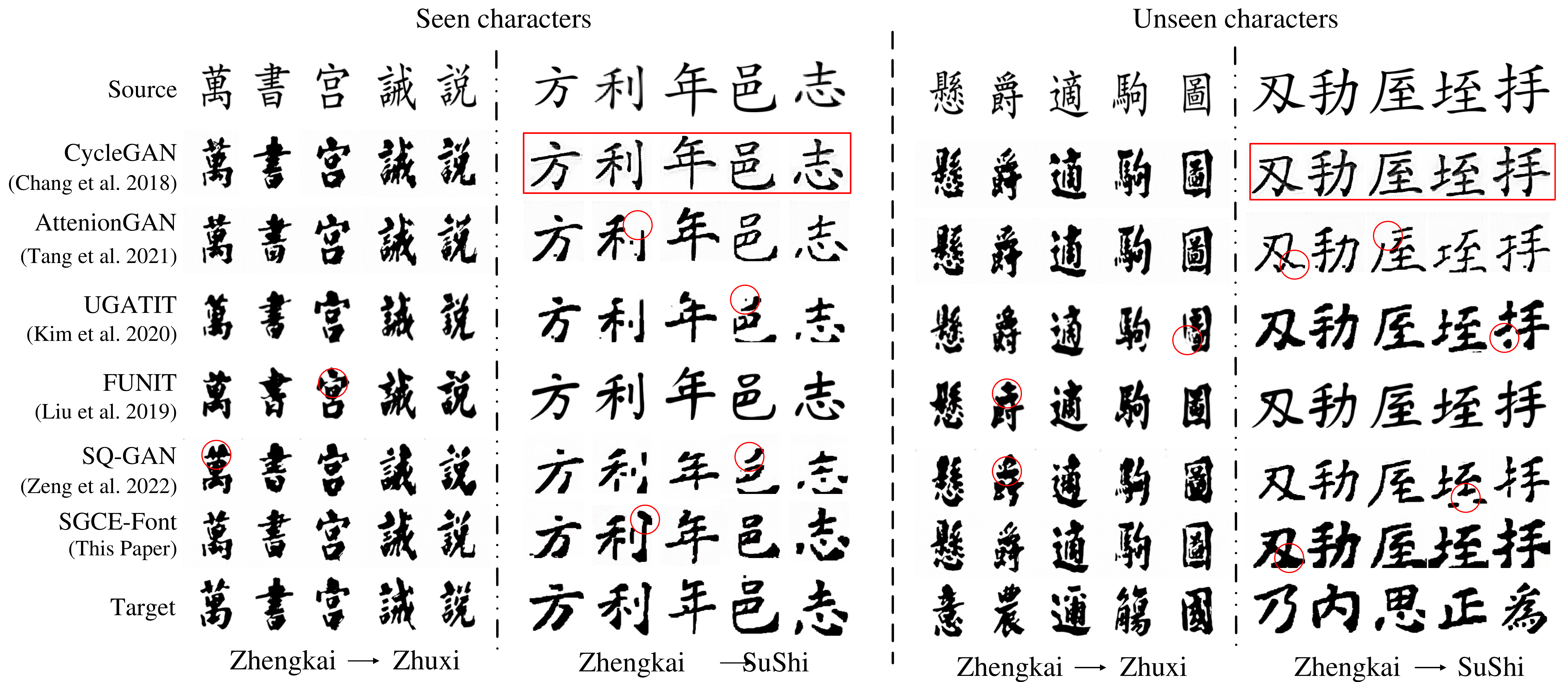}
\end{minipage}  
\hfill
\caption{Comparison on generated characters of the proposed SGCE-Font model and five baselines over two calligraphic font generation tasks and for both seen and unseen Chinese characters.}
\label{fig:Calligraphic fonts}
\end{figure*}

\begin{table}
\centering
\setlength{\tabcolsep}{0.7mm}{
\begin{tabular}{c|c|c|c|c|c|c}
\hline
& model        & hc$\rightarrow$ls  & ls$\rightarrow$hc  & zk$\rightarrow$ss   & zk$\rightarrow$zx   & avg      \\
\hline
\multirow{6}{*}{\rotatebox{90}{FID}}  & CycleGAN     & \textbf{30.173} & \textcolor{blue}{33.015} & \textcolor{blue}{204.112} & 115.039 & \textcolor{blue}{95.585}   \\
& SQ-GAN       & 50.451 & 71.220 & 295.611 & 87.388  & 126.168  \\
& FUNIT        & 54.917 & 55.576 & 213.107 & \textcolor{blue}{81.275}  & 101.219  \\
& UGATIT       & 52.689 & 48.358 & 225.062 & 224.546 & 137.664  \\
& AttentionGAN & 82.824 & 83.822 & 228.433 & 92.930  & 122.002  \\
& SGCE-Font         & \textcolor{blue}{30.695} & \textbf{30.291} & \textbf{185.330} & \textbf{70.030}  & \textbf{79.087}   \\
\hline
\multirow{6}{*}{\rotatebox{90}{MSE}}  & CycleGAN     & \textcolor{blue}{0.148}  &\textbf{0.155}  & 0.219   & 0.097   & \textcolor{blue}{0.154}    \\
& SQ-GAN       & 0.206  & 0.176  & 0.249   & \textcolor{blue}{0.095}   & 0.181    \\
& FUNIT        & 0.170  & 0.180  & 0.234   & \textcolor{blue}{0.095}   & 0.170    \\
& UGATIT       & 0.160  & 0.198  & \textbf{0.211}   & 0.179   & 0.187    \\
& AttentionGAN & 0.205  & 0.203  & 0.232   & \textbf{0.094}   & 0.184    \\
& SGCE-Font         & \textbf{0.144}  & \textcolor{blue}{0.163}  & \textcolor{blue}{0.212}   & \textbf{0.094}   & \textbf{0.153}    \\
\hline
\multirow{6}{*}{\rotatebox{90}{PSNR}} & CycleGAN     & 5.648  & \textbf{8.161}  & \textcolor{blue}{6.668}   & 10.239  & 7.679    \\
& SQ-GAN       & \textcolor{blue}{8.364}  & 7.721  & 6.095   & 10.412  & \textcolor{blue}{8.148}    \\
& FUNIT        & 7.730  & 7.480  & 6.356   & 10.361  & 7.982    \\
& UGATIT       & 8.068  & 7.075  & 6.774   & 7.543   & 7.365    \\
& AttentionGAN & 6.911  & 6.944  & 6.380   & \textcolor{blue}{10.415}  & 7.662    \\
& SGCE-Font         & \textbf{8.489}  & \textcolor{blue}{7.953}  & \textbf{6.934}   & \textbf{10.456}  & \textbf{8.458}    \\
\hline
\multirow{6}{*}{\rotatebox{90}{SSIM}} & CycleGAN     & 0.008  & \textcolor{blue}{0.519}  & \textcolor{blue}{0.456}   & 0.683   & 0.416    \\
  & SQ-GAN       & \textcolor{blue}{0.543}  & 0.499  & \textbf{0.544}   & \textcolor{blue}{0.685}   & \textbf{0.568}    \\
  & FUNIT        & 0.533  & \textcolor{blue}{0.519}  & 0.430   & 0.684   & 0.542    \\
  & UGATIT       & 0.531  & 0.458  & 0.426   & 0.498   & 0.478    \\
  & AttentionGAN & 0.445  & 0.445  & 0.429   & 0.684   & 0.501    \\
  & SGCE-Font         & \textbf{0.578}  & \textbf{0.539}  & \textcolor{blue}{0.456}   & \textbf{0.691}   & \textcolor{blue}{0.566}   \\
  \hline
\end{tabular}
\caption{Comparison on the performance of the proposed SGCE-font model and the state-of-the-art models over four calligraphy font generation tasks. The best and second best results are marked in bold and blue color respectively.
}
\label{table:seen Calligraphic fonts}
}
\end{table}
\subsection{Comparison with State-of-the-art Models}
\label{sc:comparison-sota}
In this section, we conducted twelve printing or handwriting font generation tasks to demonstrate the effectiveness of the proposed model through comparing with the state-of-the-art models including FUNIT \cite{liu2019few}, UGATIT \cite{Kim2020UGATITUG}, and AttentionGAN \cite{tang2021attentiongan}. The comparison results between the proposed model and these existing models for the twelve standard character generation are presented in Table \ref{table:comparison-sota}.

From Table \ref{table:comparison-sota}, the proposed SGCE-Font achieves the best performance in seven of these twelve font generation tasks and performs the second best in four of these twelve font generation tasks in terms of FID, as presented in the fifth row of Table \ref{table:comparison-sota}. The average FID value of the proposed model over these fourteen standard Characters generation tasks is much better than those of state-of-the-art models, as shown in the last column of Table \ref{table:comparison-sota}. Similar claims can be concluded in terms of the other three evaluation metrics from Table \ref{table:comparison-sota}. 

Some generated characters of the proposed model as well as these three state-of-the-art models for some Chinese characters in the test set are shown in Figure \ref{fig:comparison-sota}. It can be observed from this figure that SGCE-Font can produce Chinese characters with higher quality than the concerned existing models. Specifically, FUNIT \cite{liu2019few} does not perform very well when applied to the generation tasks \{Fangsong$\rightarrow$Zhengkai, Zhengkai$\rightarrow$Hupo\}, that is, the styles of generated fonts seem closer to the styles of source fonts than that of target fonts, as shown in the second row of Figure \ref{fig:comparison-sota}. 
Besides these, there are some flaws (say, missing some strokes) in the generated characters of existing models for some font generation tasks such as \{Zhengkai$\rightarrow$Handwriting\} and \{Zhengkai$\rightarrow$Shuti\}, while these flaws can be remarkably reduced by the suggested SGCE-Font model. These demonstrate the effectiveness of the proposed model.

We also present some generated characters of the proposed model and these state-of-the-art models for some unseen Chinese characters, i.e., excluded by the concerned datasets, in Figure \ref{fig:comparison-sota-unseenfont}. It can be observed that the proposed SGCE-Font model can generate more realistic and better Chinese characters than existing models for these unseen Chinese characters. This shows that the proposed model has the better generalization performance. 

\subsection{Comparison in Calligraphy Font Generation}

Besides the previous twelve printing or handwriting font generation tasks, we particularly considered four calligraphy font generation tasks to show the effectiveness of the proposed model. As presented in \ref{table:fontsize}, the total sizes of the Zhuxi (zx) and SuShi (ss) are only 920 and 166, which are much fewer than those of printing or handwriting font datasets. The quantitative comparison results over these four font generation tasks are presented in Table \ref{table:seen Calligraphic fonts}.

\begin{table*}[!htbp]
\centering\renewcommand\arraystretch{1.3}
\setlength{\tabcolsep}{1.5mm}{
\begin{tabular}{c|c|c|c|c|c|c|c|c|c|c|c|c|c}
\hline
\multicolumn{1}{c}{}& &task  & zk$\rightarrow$fs  & zk$\rightarrow$st   & fs$\rightarrow$ht  & zk$\rightarrow$hw  & fs$\rightarrow$zk   & st$\rightarrow$zk   & ht$\rightarrow$fs  & hp$\rightarrow$zk & hc$\rightarrow$ls  &zk$\rightarrow$ss &zk$\rightarrow$zx   \\
\hline\hline
\multirow{8}{*}{\rotatebox{90}{UAGTIT}}  
&\multirow{2}{*}{\rotatebox{90}{FID}}  
& Orig        & 42.824 & 118.105 & 37.918 & 77.100 & 34.032  & 128.105 & 28.324 & 76.551& 48.358 & 225.062 & 224.546   \\
& & +SGCE        & \textbf{29.516} & \textbf{33.060}  & \textbf{31.177} & \textbf{71.696} & \textbf{32.290}  & \textbf{48.030}  & \textbf{25.291} & \textbf{67.203}  & \textbf{37.943} & \textbf{103.618} & \textbf{84.286}  \\
                   
\cline{2-14}
&\multirow{2}{*}{\rotatebox{90}{MSE}}
& Orig        & \textbf{0.113}  & 0.151   & 0.170  & 0.129  & 0.090   & \textbf{0.127}   & 0.140  & 0.144  & 0.198  & \textbf{0.211}   & 0.179  \\
& & +SGCE       & 0.118  & \textbf{0.149}   & \textbf{0.144}  & \textbf{0.098}  & \textbf{0.089}   & 0.159   & \textbf{0.123}  & \textbf{0.129}    & \textbf{0.167}  & 0.249   & \textbf{0.085}  \\
                       
\cline{2-14}
&\multirow{2}{*}{\rotatebox{90}{PSNR}}
& Orig       & 9.628  & 8.336   & 7.799  & 9.342  & 10.649  &\textbf{ 9.101}   & 8.657  & 8.481  & 7.075  & \textbf{6.774}   & 7.543    \\
& & +SGCE       & \textbf{9.479}  & \textbf{8.398}   & \textbf{8.571}  & \textbf{10.214} & \textbf{10.709}  & 8.039   & \textbf{9.213}  & \textbf{8.981}   & \textbf{7.855}  & 6.086   & \textbf{10.863}  \\
                       
\cline{2-14}
&\multirow{2}{*}{\rotatebox{90}{SSIM}}
& Orig        & \textbf{0.549}  & 0.528   & 0.549  & 0.271  & 0.589   & 0.486   & 0.464  & 0.451   & 0.458  & 0.426   & 0.498    \\
& & +SGCE        & 0.548  & \textbf{0.613}   & \textbf{0.584}  & \textbf{0.322}  & \textbf{0.594}   & \textbf{0.522}   & \textbf{0.602}  & \textbf{0.472}  & \textbf{0.505}  & \textbf{0.433}   & \textbf{0.704}    \\
\hline\hline

\multirow{8}{*}{\rotatebox{90}{AtteniongGAN}}  
&\multirow{2}{*}{\rotatebox{90}{FID}}  
& Orig  & 32.319 & 92.032  & 32.796 & 26.861 & 57.464  & 51.844  & 37.740 & \textbf{55.689}   & 83.822 & \textbf{228.433} & 92.930 \\
& & +SGCE & \textbf{19.163} & \textbf{43.024}  & \textbf{28.997} & \textbf{20.053} & \textbf{46.997}  & \textbf{38.718}  & \textbf{25.601} & 58.478 & \textbf{24.550} & 237.649 & \textbf{72.509}   \\
\cline{2-14}
& \multirow{2}{*}{\rotatebox{90}{MSE}}
& Orig  & 0.109  & 0.197   & 0.170  & 0.097  & \textbf{0.095}   & 0.151   & 0.125  & 0.143  & 0.203  & 0.232   & \textbf{0.094}    \\
& & +SGCE & \textbf{0.106}  & \textbf{0.151}   & \textbf{0.105}  & \textbf{0.094}  & \textbf{0.095}   & \textbf{0.122}   & \textbf{0.105}  & \textbf{0.134}   & \textbf{0.143}  & \textbf{0.220}   & \textbf{0.094}    \\
\cline{2-14}
& \multirow{2}{*}{\rotatebox{90}{PSNR}}
& Orig  & 9.826  & 7.084   & 7.749  & 10.241 & \textbf{10.398}  & 8.265   & 9.148  & 8.470  & 6.944  & 6.380   & 10.415    \\
& & +SGCE & \textbf{9.941}  & \textbf{8.334}   & \textbf{9.998}  & \textbf{10.391} & 10.368  & \textbf{9.301}   & \textbf{9.880}  & \textbf{8.780}  & \textbf{8.509}  & \textbf{6.744}   & \textbf{10.450}    \\
\cline{2-14}
& \multirow{2}{*}{\rotatebox{90}{SSIM}}
& Orig  & 0.580  & 0.451   & 0.463  & 0.332  & \textbf{0.588}   & 0.433   & 0.507  & 0.452  & 0.445  & 0.429   & 0.684     \\
& & +SGCE & \textbf{0.589}  & \textbf{0.520}   & \textbf{0.571}  & \textbf{0.344}  & 0.587   & \textbf{0.506}   & \textbf{0.602}  & \textbf{0.475}  & \textbf{0.562}  & \textbf{0.444}   & \textbf{0.689}     \\
\hline\hline

\multirow{8}{*}{\rotatebox{90}{SQ-GAN}}  
& \multirow{2}{*}{\rotatebox{90}{FID}}   
& Orig        & 49.865 & 71.694  & 38.896 & 49.532 & 99.038  & 146.293 & \textbf{37.546} & 305.931 & 50.451 & 295.611 & 115.039  \\
& & +SGCE       & \textbf{45.790} & \textbf{49.004}  & \textbf{35.345} & \textbf{46.152} & \textbf{40.964}  & \textbf{101.882} & 84.855 & \textbf{88.694}& \textbf{30.805} & \textbf{125.620} & \textbf{77.896}    \\
                       
\cline{2-14}
& \multirow{2}{*}{\rotatebox{90}{MSE}}
& Orig & 0.147  & 0.199   & 0.186  & 0.106  & 0.152   & \textbf{0.158}   & 0.143  & 0.161  & 0.206  & \textbf{0.249}   & \textbf{0.090}    \\
& & +SGCE       & \textbf{0.116}  & \textbf{0.187}   & \textbf{0.123}  & \textbf{0.100}  & \textbf{0.123}   & 0.164   & \textbf{0.133}  & \textbf{0.147}    & \textbf{0.164}  & 0.262   & 0.094  \\
                       
\cline{2-14}
& \multirow{2}{*}{\rotatebox{90}{PSNR}}
& Orig         & \textbf{9.822}  & 7.058   & 7.817  & 9.791  & \textbf{9.369}   & \textbf{8.050}   & 8.626  & 7.404    & \textbf{8.364}  & \textbf{6.095}   & \textbf{10.635}   \\
& & +SGCE       & 9.539  & \textbf{7.323}   & \textbf{9.318}  & \textbf{10.082} & 9.180   & 7.887   & \textbf{8.852}  & \textbf{8.378}    & 7.891  & 5.858   & 10.432  \\
                       
\cline{2-14}
& \multirow{2}{*}{\rotatebox{90}{SSIM}}
& Orig        & \textbf{0.570}  & 0.448   & 0.478  & 0.293  & 0.530   & \textbf{0.428}   & 0.485  & 0.359  & \textbf{0.543}  & \textbf{0.544}   & \textbf{0.754}     \\
& & +SGCE       & \textbf{0.570}  & \textbf{0.488}   & \textbf{0.583}  & \textbf{0.320}  & \textbf{0.535}   & 0.423   & \textbf{0.497}  & \textbf{0.465} & 0.532  & 0.405   & 0.690    \\
\hline\hline

\multirow{8}{*}{\rotatebox{90}{FUNIT}} 
& \multirow{2}{*}{\rotatebox{90}{FID}}  
& Orig       & \textbf{37.505} & 43.675  & 64.990 & \textbf{75.553} & 151.728 & 207.221 & \textbf{40.677} & 254.404  & 55.576 & 282.494 & 139.837 \\
& & +SGCE    & 38.683 & \textbf{41.299}  & \textbf{57.760} & 77.190 & \textbf{133.800} & \textbf{205.078} & 53.910 & \textbf{132.696} & \textbf{55.480} & \textbf{111.208} & \textbf{101.170}  \\
                      
\cline{2-14}
& \multirow{2}{*}{\rotatebox{90}{MSE}}
& Orig    & \textbf{0.127}  & 0.224   & 0.212  & \textbf{0.105}  & 0.138   & 0.180   & 0.178  & \textbf{0.188}   & \textbf{0.180}  & \textbf{0.167}   &\textbf{0.093}    \\
& & +SGCE      & 0.128  & \textbf{0.220}   & \textbf{0.212}  & 0.106  & \textbf{0.137}   & \textbf{0.178}   & \textbf{0.177}  & 0.189   & 0.183  & 0.182   & 0.095     \\
                      
\cline{2-14}
& \multirow{2}{*}{\rotatebox{90}{PSNR}}
& Orig        & \textbf{9.075}  & 6.523   & 6.772  & \textbf{9.861}  & 8.660   & 7.476   & 7.525  & \textbf{7.276}   & \textbf{7.480}  & \textbf{7.819}   & \textbf{10.414}  \\
& & +SGCE       & 9.042  &\textbf{ 6.604 }  & \textbf{6.777}  & 9.791  & \textbf{8.727}   & \textbf{7.519}   & \textbf{7.553}  & 7.271   & 7.407  & 7.476   & 10.287   \\
                      
\cline{2-14}
& \multirow{2}{*}{\rotatebox{90}{SSIM}}
& Orig        & \textbf{0.525}  & 0.426   & \textbf{0.388}  & \textbf{0.295}  & 0.485   & 0.412   & 0.389  & \textbf{0.386}  & \textbf{0.519}  & 0.397   & 0.686       \\
& & +SGCE        & 0.522  & \textbf{0.433}   & 0.386  & 0.291  & \textbf{0.498}   & \textbf{0.416}   & \textbf{0.395}  & 0.380  & \textbf{0.519}  & \textbf{0.498}   & \textbf{0.691}    \\
\hline
\end{tabular}
\caption{Comparison on the performance of the state-of-the-art models and their refined models equipped with the suggested SGCE module. For each model, ``Orig'' and ``+SGCE'' represent the concernced model and its refined model equipped with SGCE, respectively. }
\label{table:sota+sgce}
}
\end{table*}
\begin{table*}[!htbp]
\centering\renewcommand\arraystretch{1.3}
\begin{tabular}{c|c|c|c|c|c|c|c|c|c|c}
\hline
\multicolumn{1}{c}{}& &task  & zk$\rightarrow$fs  & zk$\rightarrow$st   & fs$\rightarrow$ht  & zk$\rightarrow$hw  & fs$\rightarrow$zk   & st$\rightarrow$zk   & ht$\rightarrow$fs  & hp$\rightarrow$zk   \\
\hline

\multirow{8}{*}{\rotatebox{90}{StrokeGAN}}  
& \multirow{2}{*}{\rotatebox{90}{FID}}  
& Orig     & 57.851 & 71.302  & 28.972 & 37.779 & 67.507  & 71.302  & 39.610 & 140.227   \\
& & +SGCE    & \textbf{33.283} & \textbf{25.230}  & \textbf{15.861} & \textbf{28.201} & \textbf{24.959}  & \textbf{44.303}  & \textbf{21.837} & \textbf{89.440}    \\
\cline{2-11}
& \multirow{2}{*}{\rotatebox{90}{MSE}}
& Orig    & 0.120  & \textbf{0.158}   & 0.188  & 0.104  & \textbf{0.104}   & \textbf{0.125}   & 0.129  & 0.181      \\
& & +SGCE    & \textbf{0.114}  & 0.189   & \textbf{0.123}  & \textbf{0.095}  & 0.128   & 0.169   & \textbf{0.117}  & \textbf{0.168}      \\
\cline{2-11}
& \multirow{2}{*}{\rotatebox{90}{PSNR}}
& Orig     & 9.359  & 6.774   & 7.318  & 9.906  & \textbf{10.002}  & 6.774   & 9.043  & 8.379      \\
& & +SGCE     & \textbf{9.613}  & \textbf{7.311}   & \textbf{9.362}  & \textbf{10.336} & 8.996   & \textbf{7.763}   & \textbf{9.495}  & \textbf{9.010}     \\
\cline{2-11}
& \multirow{2}{*}{\rotatebox{90}{SSIM}}
& Orig     & 0.553  & 0.464   & 0.462  & 0.301  & \textbf{0.574}   & \textbf{0.464}   & 0.533  & 0.448     \\
& & +SGCE    & \textbf{0.577}  & \textbf{0.495}   & \textbf{0.600}  & \textbf{0.448}  & 0.520   & 0.432   & \textbf{0.567}  & \textbf{0.503}      \\
\hline
\end{tabular}
\caption{Comparison on the performance of StrokeGAN and its refined model equipped with the suggested SGCE module. }
\label{table:StorkeGAN+sgce}
\end{table*}

\begin{figure*}[htbp]
\begin{minipage}[b]{0.99\linewidth}
\centering
\includegraphics*[scale=0.25]{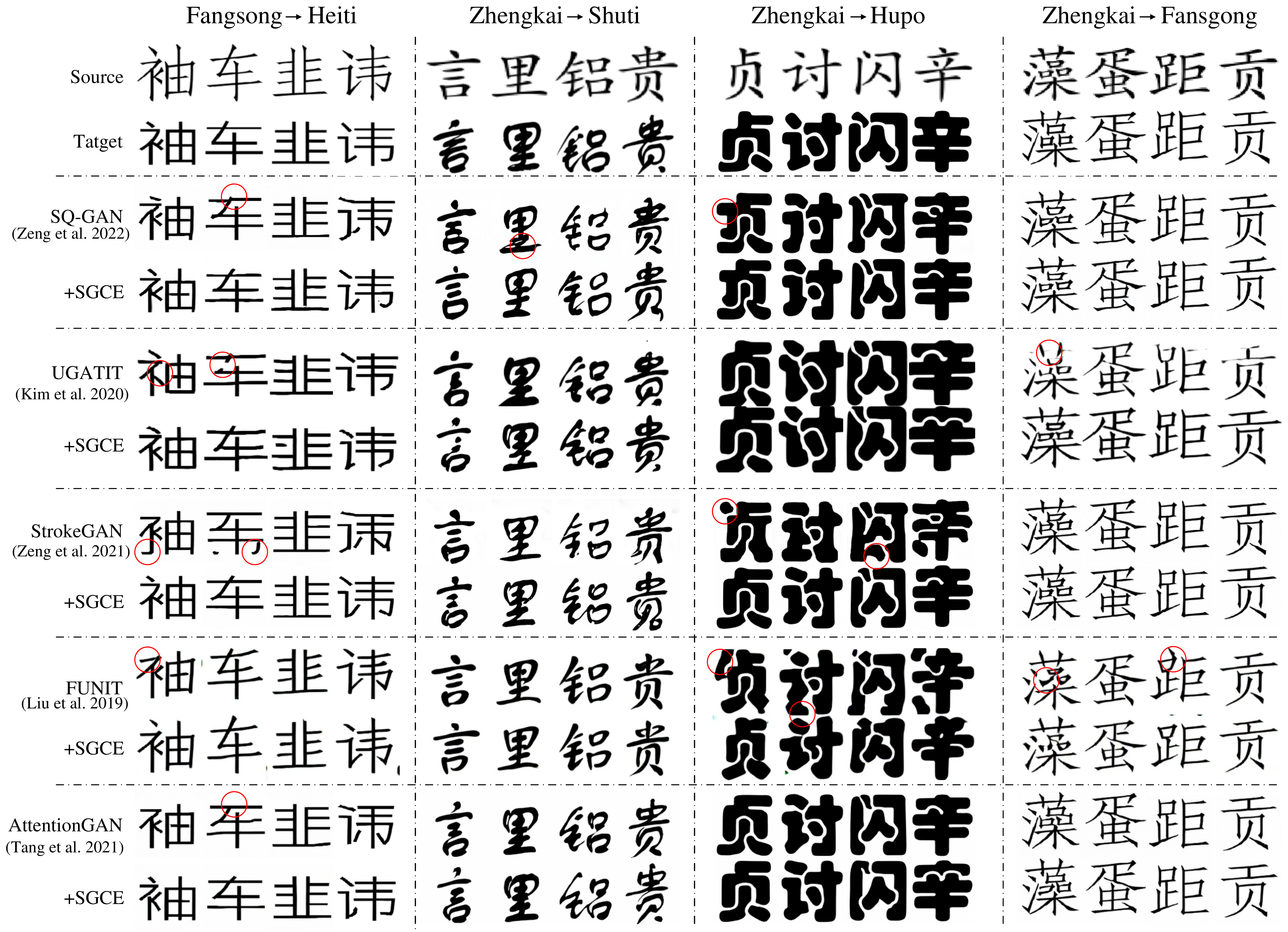}
\end{minipage}
\hfill
\caption{Generated characters of the state-of-the-art models as well as their refined models equipped with the proposed SGCE module.
}
\label{fig:generalization-characters}
\end{figure*}

From Table \ref{table:seen Calligraphic fonts}, the proposed SGCE-Font achieves the best performance in average in terms of FID, MSE and PSNR, while acheives the second best performance in average in terms of SSIM. Specifically, in terms of FID, the proposed SGCE-Font model achieves the best results in the font generation tasks \{ls$\rightarrow$hc, zk$\rightarrow$ss, zk$\rightarrow$zx\}, while achieves the second best result in the font generation tasks \{hc$\rightarrow$ls\}. Similar claims can be also claimed in terms of the other three evaluation metrics. 
We also present some visualization comparisons between the proposed model and five state-of-the-art models over two calligraphy font generation tasks. The comparison results for the characters in the test set and the unseen characters excluded by the datasets are presented in Figure \ref{fig:Calligraphic fonts}. It can be observed from Figure \ref{fig:Calligraphic fonts} that the quality of generated characters of the proposed model is better than the baselines.
These show clearly the effectiveness of the proposed model in these calligraphy font generation tasks.
\setlength{\parskip}{0pt}
\subsection{Generalization of Proposed SGCE Module}
\label{sc:generalization}

As pointed out before, the proposed SGCE module can be used as a plug-and-play module for the Chinese font generation models. In the following, we adapted the proposed SGCE module to five existing models including UGATIT \cite{Kim2020UGATITUG}, AttentionGAN \cite{tang2021attentiongan}, StrokeGAN \cite{zeng2021strokegan}, SQ-GAN \cite{zeng2022SQ-GAN} and FUNIT \cite{liu2019few} to further enhance their performance. We incorporated SGCE into each base model by a similar way in Figure \ref{fig:SGCE-Font}. For UGATIT, AttentionGAN, SQ-GAN and FUNIT, we implemented eleven font generation tasks including eight printing or handwriting font generation tasks and three calligraphy font generation tasks to evaluate the effectiveness of the proposed SCGE module, while for StrokeGAN, we did not tested its performance in these three calligraphy font generation tasks since it was expensive to build up the stroke encodings of the traditional Chinese characters involved in the calligraphy font generation tasks.

The quantitative comparison results are presented in Table \ref{table:sota+sgce} and Table \ref{table:StorkeGAN+sgce}. From Table \ref{table:sota+sgce} and Table \ref{table:StorkeGAN+sgce}, the performance of these five concerned models can be substantially improved over most of font generation tasks, through equipping with the proposed SGCE module in terms of these four evaluation metrics. We also present some visualization comparison results in Figure \ref{fig:generalization-characters}. It can be observed from Figure \ref{fig:generalization-characters} that the quality of generated characters can be improved by equipping with the proposed SGCE module. These show clearly the effectiveness of the proposed SGCE module.

\section{Conclusion}
\label{sc:conclusion}
Using what kind of information as the guidance information and how to use it to reduce the mode collapse issue are two important questions for the predominated GAN based models for the Chinese font generation. This paper proposed an effective guidance module called SGCE for the Chinese font generation models, where the skeleton information is adopted and used through a channel expansion way, motivated by the observation that the skeleton embodies both local and global information of a Chinese character. The channel expansion way of using skeleton directly imposes the skeleton guidance information onto the generator, which is more effective than the existing way that imposes the guidance information onto the discriminator. 
Extensive experiments are conducted to demonstrate the effectiveness of the proposed SGCE module. Experimental results show that the proposed SGCE module is more effective in reducing the mode collapse issue suffered by the well-known CycleGAN, and can be easily adapted to existing models as a plug-and-play module to further improve their performance. One future direction is to adapt the proposed module to the generation of other languages (e.g., Korean) or other scenarios like the few-shot Chinese font generation.

\bibliographystyle{ieeetr}
\bibliography{SGCEFont}
\end{document}